\documentclass[letterpaper]{article} % DO NOT CHANGE THIS
\usepackage{aaai2026}  % DO NOT CHANGE THIS
\usepackage{times}  % DO NOT CHANGE THIS
\usepackage{helvet}  % DO NOT CHANGE THIS
\usepackage{courier}  % DO NOT CHANGE THIS
\usepackage[hyphens]{url}  % DO NOT CHANGE THIS
\usepackage{graphicx} % DO NOT CHANGE THIS
\usepackage{natbib}  % DO NOT CHANGE THIS AND DO NOT ADD ANY OPTIONS TO IT
\usepackage{caption} % DO NOT CHANGE THIS AND DO NOT ADD ANY OPTIONS TO IT
\usepackage{colortbl}  % 这里引入 colortbl 宏包，支持表格颜色
\usepackage{multirow}  % 如果你用到了 \multirow，也建议引入
\usepackage{booktabs}  % 如果你用到了 \toprule \midrule \bottomrule 这些表格美化命令
\usepackage{makecell}
\usepackage{tabularx}
\usepackage{pifont}   % 用于 \times
\usepackage{times}  % DO NOT CHANGE THIS
\usepackage{helvet}  % DO NOT CHANGE THIS
\usepackage{courier}  % DO NOT CHANGE THIS
\usepackage{amsmath} 
\usepackage{amssymb}   % For symbols 
\usepackage{subcaption} % 支持subfigure组合
\usepackage[table]{xcolor}
\usepackage{float}
\definecolor{lightblue}{RGB}{204, 231, 242} 
\usepackage{algorithm}
\usepackage{algorithmic}

\usepackage{newfloat}
\usepackage{listings}
\DeclareCaptionStyle{ruled}{labelfont=normalfont,labelsep=colon,strut=off} % DO NOT CHANGE THIS
\floatstyle{ruled}
\newfloat{listing}{tb}{lst}{}
\floatname{listing}{Listing}
\title{EfficientFSL: Enhancing Few-Shot Classification via Query-Only Tuning in Vision Transformers}
\author{
    Wenwen Liao\textsuperscript{\rm 1},
    Hang Ruan\textsuperscript{\rm 1},
    Jianbo Yu\textsuperscript{\rm 2}\thanks{Corresponding to jb\_yu@fudan.edu.cn},
    Bing Song\textsuperscript{\rm 3},
    Yuansong Wang\textsuperscript{\rm 4},
    Xiaofeng Yang\textsuperscript{\rm 2}
}
\affiliations{
    \textsuperscript{\rm 1}College of Intelligent Robotics and Advance Manufacturing, Fudan University, Shanghai China\\
    \textsuperscript{\rm 2}School of Microelectronics, Fudan University, Shanghai China\\
    \textsuperscript{\rm 3}School of Information Science and Engineering, East China University of Science and Technology, Shanghai China\\
    \textsuperscript{\rm 4}Tsinghua Shenzhen International Graduate School, Tsinghua University, Shenzhen China
}

\usepackage{bibentry}
\begin{document}

\maketitle

\begin{abstract}
Large models such as Vision Transformers (ViTs) have demonstrated remarkable superiority over smaller architectures like ResNet in few-shot classification, owing to their powerful representational capacity. However, fine-tuning such large models demands extensive GPU memory and prolonged training time, making them impractical for many real-world low-resource scenarios. To bridge this gap, we propose \textbf{EfficientFSL}, a query-only fine-tuning framework tailored specifically for few-shot classification with ViT, which achieves competitive performance while significantly reducing computational overhead. EfficientFSL fully leverages the knowledge embedded in the pre-trained model and its strong comprehension ability, achieving high classification accuracy with an extremely small number of tunable parameters.  
Specifically, we introduce a lightweight trainable Forward Block to synthesize task-specific queries that extract informative features from the intermediate representations of the pre-trained model in a query-only manner. We further propose a Combine Block to fuse multi-layer outputs, enhancing the depth and robustness of feature representations. Finally, a Support-Query Attention Block mitigates distribution shift by adjusting prototypes to align with the query set distribution. 
With minimal trainable parameters, EfficientFSL achieves state-of-the-art performance on four in-domain few-shot datasets and six cross-domain datasets, demonstrating its effectiveness in real-world applications.
\end{abstract}

\begin{links}
    % \link{Code}{https://github.com/WenwenLiaoo/AAAI26-EfficientFSL}
    
    % \link{Datasets}{https://aaai.org/example/datasets}
    % \link{Extended version}{https://aaai.org/example/extended-version}
\end{links}

\section{Introduction}

Despite the remarkable progress of deep learning in computer vision, it typically relies on large amounts of labeled data. In contrast, Few-shot Learning (FSL) \cite{finn2017model,munkhdalai2018rapid,antoniou2018train} enables the construction of models that can recognize and classify previously unseen categories using only a very limited number of samples. Since the model must learn discriminative features from minimal labeled data \cite{zhang2022deepemd,afrasiyabi2022matching}, the representation capability of the model becomes crucial.

To enhance this capability, many recent approaches \cite{li2024knn,zhang2024simple,li2024fewvs} have replaced CNN-based ResNet models with Vision Transformers (ViT) \cite{dosovitskiy2020image} for feature extraction, achieving significant performance gains. A common strategy involves pre-training on a general, large-scale dataset, followed by adaptation to few-shot tasks. The most straightforward way to adapt is to fully fine-tune all parameters of the pre-trained model. However, this requires storing a complete set of model parameters for each individual task, leading to considerable storage overhead \cite{zhai2022scaling}.

To improve storage efficiency, researchers have explored Parameter-Efficient Transfer Learning (PETL) methods \cite{chen2022adaptformer,hu2022lora,jia2022visual,jie2023fact,jie2023revisiting}, which adapt large pre-trained models to downstream tasks by tuning only a small portion of parameters. Typically, these methods insert small modules into the frozen pre-trained model and only update the parameters of these added components. However, these methods inevitably modify the feature flow and remain coupled with the backbone weights, risking overfitting and reduced generalization, especially in data-limited settings like FSL. To overcome this, we propose a novel query-only paradigm \textbf{EfficientFSL}, a fine-tuning method tailored specifically for few-shot classification with ViT. The core distinction from existing methods lies in EfficientFSL’s approach: instead of modifying the backbone, it freezes it entirely and introduces a lightweight querying module to selectively extract task-relevant information.

As illustrated in Figure \ref{fig:pipeline}, EfficientFSL takes the intermediate representations from a pre-trained model as input and integrates them with task-specific knowledge through a stack of modular components. First, EfficientFSL follows a decoupling design, employing a lightweight Forward Block that consists of a trainable Active Block and a Frozen Block. The Active Block focuses on learning task-specific knowledge by generating adaptive queries. These queries guide the model to attend to relevant information. The following Frozen Block provides general knowledge by reusing and freezing the intermediate representations from the pre-trained backbone, effectively preserving prior knowledge while avoiding unnecessary parameter updates. Next, building on the rich hierarchical features from the Frozen Block, we introduce the Combine Block, which adaptively aggregates multi-layer features using the final layer's output as guidance. Throughout the architecture, Bottleneck-Structured Fully Connected Layers are utilized to reduce the number of trainable parameters.

Importantly, to address the distribution shift between the support set and the query set, Support-Query Attention Block (SQ Attention Block) is proposed. This component adjusts the position of prototypes to better align with the center of the corresponding query distribution. EfficientFSL outperforms recent approaches across all benchmarks.

The contributions are summarized as follows:
\begin{itemize}
  \item \textbf{A Novel PEFT Framework for FSL.} We propose \textbf{EfficientFSL}, a parameter-efficient fine-tuning framework for few-shot classification. It effectively integrates pre-trained representations with task-specific knowledge while significantly reducing the number of trainable parameters.
  \item \textbf{Enhance Support-Query Alignment.} To mitigate the distribution shift between the support and query sets and improve generalization in few-shot scenarios, we propose the SQ Attention Block, which dynamically adjusts prototype positions by aligning them with the query set distribution.
  \item \textbf{Outstanding Performance.} Extensive experiments on 4 in-domain few-shot datasets and 6 cross-domain datasets, validate the effectiveness of our method and its individual components.
\end{itemize}

\section{Related Works}

\subsection{Few-Shot Learning}
Few-shot learning (FSL) aims to learn classifiers for new categories using only a few labeled examples. Studies on the few-shot classification problem can be roughly divided into the following categories: (i) Optimization-based methods leverage meta-learning to acquire well-initialized parameters that enable adaptation to new classes with minimal optimization steps \cite{finn2017model, Ravi2017, Rusu2019,Baik2020}. (ii) Data augmentation-based approaches adaptively predict classifier weights for new categories based on the feature embeddings of new examples \cite{Wang2018, Zhang2019, Wang2020}. (iii) Metric-based methods focus on learning a metric space in which a query example can be classified using a nearest neighbor approach \cite{Snell2017, Vinyals2016, Koch2015, Chen2020}. 

Among these, metric-based methods have emerged as particularly appealing due to their simplicity, efficiency in low-data regimes. These approaches aim to learn an embedding space where samples from the same category are close together, while samples from different categories are farther apart \cite{Fu2022}. Typically, this involves projecting all input instances into vectors of fixed dimensionality, with predictions for query samples made using methods such as nearest neighbor classifiers \cite{Vinyals2016}, robust category weights \cite{Tang2022}, parameterized metrics \cite{Sung2018}, or other task-specific distance metrics \cite{Liu2020}.

 % \cite{Ji2020, Ding2020, Pahde2021, Gogoi2022}. PN exemplifies the core idea of metric-based FSL by calculating the mean vector of embedded examples for each class, which serves as a prototype for classification. In this work, we build upon the Prototypical Network framework to enhance few-shot classification performance.

\subsection{Parameter-Efficient Transfer Learning}

Updating all parameters of large-scale pre-trained models can lead to substantial computational overhead, thereby hindering the efficiency of rapid model adaptation and deployment. To address this issue, researchers have proposed Parameter-Efficient Transfer Learning (PETL), which enables effective adaptation to downstream tasks by updating only a small subset of parameters while keeping the majority of the model frozen.

When applied to large vision models such as ViT, PETL offers an efficient and cost-effective solution. Common PETL techniques include the use of soft prompts~\cite{li2021prefix, jia2022visual}, lightweight adapter modules~\cite{houlsby2019parameter, chen2022adaptformer, he2021towards}, and low-rank matrix decomposition~\cite{zhang2023adalora, valipour2022dylora, hu2022lora}. Notably, recent studies have demonstrated that in few-shot settings with large language models, parameter-efficient fine-tuning (PEFT) can achieve performance comparable to or even surpassing that of full fine-tuning~\cite{liu2022few, yang2024mixture}. This reveals a natural synergy between PETL strategies and FSL.

\begin{figure*}[ht]
  \centering
  \includegraphics[width=\linewidth]{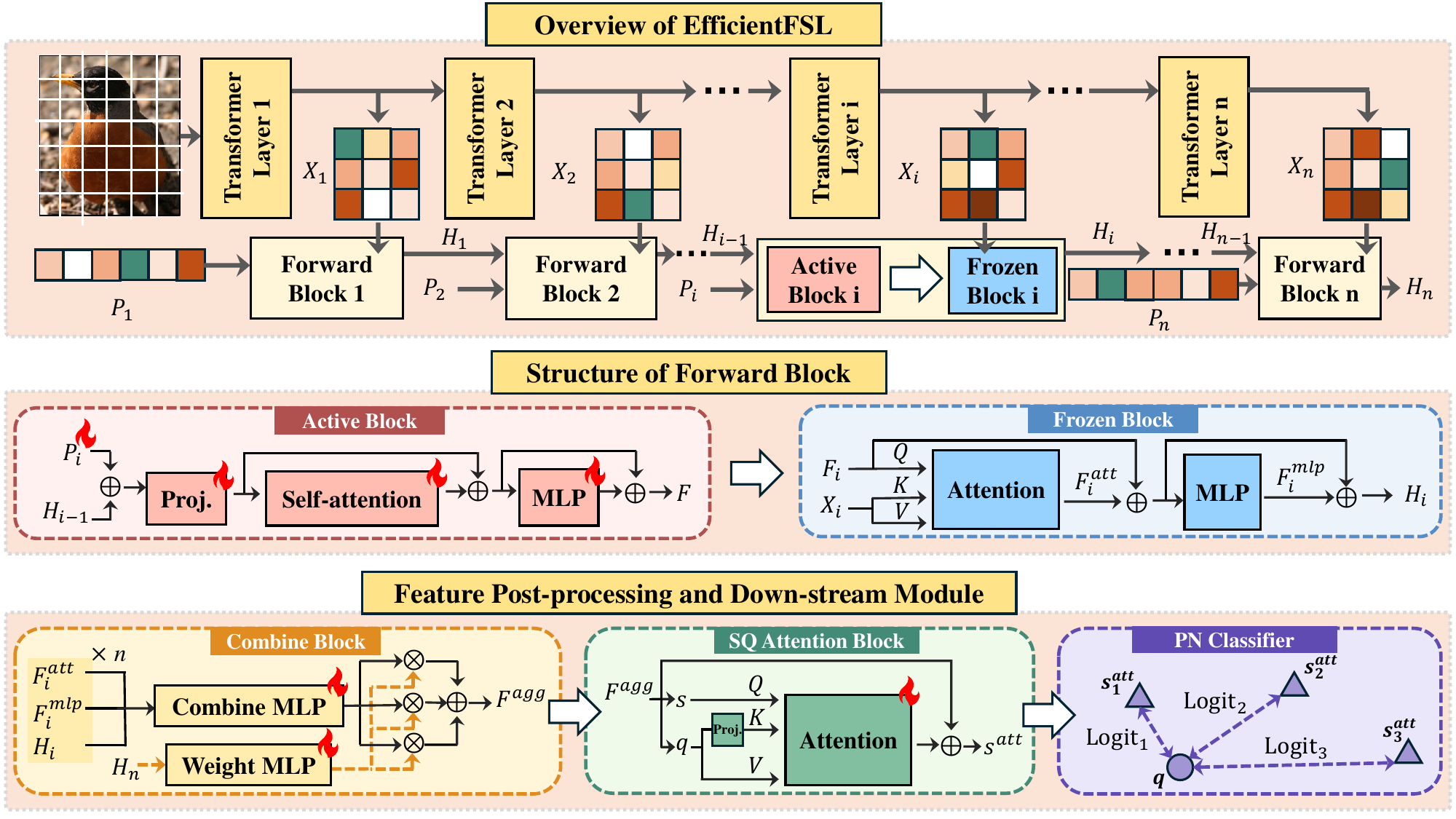} % 调整宽度
  \caption{An overview of our pipeline. EfficientFSL takes intermediate representations from a pre-trained model as input and integrates task-specific knowledge through the Forward Block, Combine Block, and SQ Attention Block, followed by classification using a PN classifier.}
  \label{fig:pipeline}
\end{figure*}

\section{Preliminaries}
\label{sec:intro}

\subsection{Problem Setting}
In few-shot classification, the dataset is divided into a train set \( \mathcal{D}^{\textit{train}} \), and a test set, \( \mathcal{D}^{\textit{test}} \), where there is no overlap between the classes of the train and test sets. Each set is further split into two subsets. One is support set \( \mathcal{S} = \{(\boldsymbol{x}_i, y_i)\}_{i=0}^{N \times K} \), containing a few labeled examples for each class used for forming the prototypes of categories. The other subset, the query set \( \mathcal{Q} = \{(\boldsymbol{x}_i, y_i)\}_{i=0}^{N \times Q} \), consists of unlabeled examples used to compute their distances to the prototypes for classification. Here, \( N \) denotes the number of classes per episode, and \( K \) represents the number of support examples per class, and \( Q \) the number of query examples per class. This setting is commonly known as the \textit{N}-way \textit{K}-shot classification problem, where \( \boldsymbol{x}_i \) is the input image and \( y_i \) its label.

\subsection{Prototypical Network}

A representative and widely adopted metric-based method is the Prototypical Network (PN) \cite{Snell2017}, which has gained popularity for its ability to learn discriminative and compact class representations in few-shot classification tasks.

We adopt PN as the classification head. The feature extractor \( \textit{f} \) is trained to construct an embedding space where samples from the same class are mapped close to one another, while those from different classes are pushed farther apart. For each class \( c \), the prototype \( \boldsymbol{s}^c \) is computed as the mean of the feature representations of its support samples, as defined by:

\begin{equation}
\boldsymbol{s}^c = \frac{1}{|\mathcal{D}^c|} \sum_{(\boldsymbol{x}_i, y_i) \in \mathcal{D}^c} f(\boldsymbol{x}_i),
\label{eq1}
\end{equation}

\noindent where \( \mathcal{D}^c \) denotes the set of support examples belonging to class \( c \). Classification is then performed by measuring the distance between each query sample and the prototypes of all classes, enabling efficient and effective prediction.

\section{Methodology}

\subsection{Overview}

From the upper part of Figure \ref{fig:pipeline}, it can be observed that EfficientFSL takes the intermediate features from each layer of the pre-trained ViT as the input for each block in EfficientFSL.  

Each block consists of an active sub-block and a frozen sub-block:  
The Active Block is a lightweight module designed to learn dataset-specific knowledge, adapt to specific downstream tasks, and particularly enhance few-shot classification performance.  
The Frozen Block utilizes the tokens generated by the Active Block as queries, while using the intermediate features from the pre-trained ViT as keys and values. By reusing the original model's weights, it preserves the inherent capability of ViT. Afterward, a Combine Block is employed to adaptively fuse the rich features extracted from each Frozen Block using adaptive weights, resulting in a unified feature output.  

To reduce the distribution bias between the support set and query set, the SQ Attention Block is introduced, which refines the distribution of prototypes.  
Finally, PN serves as the classification head to perform classification.

\subsection{Forward Block}

\subsubsection{Active Block.}

The Active Block is a lightweight network used to adapt to downstream tasks. The output from the previous layer \( H_{i-1} \) is added with a trainable prompt \( P_i \), and then feed it into a bottleneck-structured projection layer to obtain the output \( Z_i \). These per-layer added prompts and the projection layer \( \text{Proj}(\cdot) \) enhance the Active Block's ability to capture task-specific knowledge.

\begin{equation}
Z_i = \text{Proj}(H_{i-1} + P_i).
\end{equation}

Inspired by the Transformer structure, \( Z_i \) interacts through a self-attention layer and is further passed into a Multilayer Perceptron (MLP). Notably, to maintain parameter efficiency, we extensively use bottleneck-structured fully connected layers here to reduce the number of parameters. In the attention layer \( \text{Att}(\cdot) \), \( Z_i \) passes through three bottleneck-structured projection layers to produce the query \( Q_i^A \), key \( K_i^A \), and value \( V_i^A \). Then, the computation in the attention layer is as follows, controlled by a scale factor \( \xi \):

\begin{equation}
Z_i' = \xi \cdot \text{Att}(Q_i^A, K_i^A, V_i^A) + Z_i.
\end{equation}

Subsequently, \( Z_i' \) is passed through a layer normalization \( \text{LN}(\cdot) \) and a fully connected layer \( \text{MLP}(\cdot) \) to obtain \( F_i \), controlled by another scale factor \( \zeta \):

\begin{equation}
F_i = \zeta \cdot \text{MLP}(\text{LN}(Z_i')) + Z_i'.
\end{equation}

Since the Active Block is trained from scratch on the downstream dataset, its output \( F_i \) captures task-specific knowledge. In the next section, \( F_i \) will be used in a query-only manner.
\subsubsection{Frozen Block.}

The Frozen Block reuses and freezes the pre-trained parameters for feature interaction, requiring no additional training, which ensures the efficiency of our training process. Its input consists of \( F_i \) obtained from the Active Block, as well as the outputs from each layer in the ViT forward propagation process \( \{X_1, X_2, \dots, X_n\} \), where \( n \) is the number of layers in the pre-trained ViT model. 

In the attention layer \( \text{Att}'(\cdot) \), we use \( F_i \) to obtain the query \( Q_i^F \), and \( X_i \) to obtain the key \( K_i^F \) and value \( V_i^F \). The subsequent attention computation is as follows:

\begin{equation}
F_i^{att} = \text{Att}'(Q_i^F, K_i^F, V_i^F) + F_i.
\end{equation}

Next, \( F_i^{att} \) passes through a layer normalization \( \text{LN}(\cdot) \) and a fully connected layer \( \text{MLP}'(\cdot) \) to obtain \( F_i^{mlp} \):

\begin{equation}
F_i^{mlp} = \text{MLP}'(\text{LN}(F_i^{att})).
\end{equation}

Finally, the feature \( H_i \) is obtained as:

\begin{equation}
H_i = F_i^{mlp} + F_i^{att}.
\end{equation}

To maximize the potential of the Frozen Block, in addition to the final output \( H_i \), we also leverage the features \( F_i^{att} \) and \( F_i^{mlp} \) from the attention module and the fully connected layer for classification.

\subsection{Combine Block}

\subsubsection{Feature Alignment using a Shared Combine MLP.}
Through the Forward Block, hierarchical features are obtained as follows: 

\(\left\{ F_1^{att}, F_1^{mlp}, H_1, F_2^{att}, F_2^{mlp}, H_2, \dots, F_n^{att}, F_n^{mlp}, H_n \right\}\).

First, we align all features using a shared combine MLP. Specifically, we employ a bottleneck-structured fully connected layer and share the projection layers across all features to improve parameter efficiency. The transformation for each feature is given by:

\begin{equation}
\left\{
\begin{aligned}
\hat{F}_i^{att} &= \text{MLP}_{\text{Combine}}^{\text{shared}}(F_i^{att}),  \\ 
\hat{F}_i^{mlp} &= \text{MLP}_{\text{Combine}}^{\text{shared}}(F_i^{mlp}),  \\ 
\hat{H}_i &= \text{MLP}_{\text{Combine}}^{\text{shared}}(H_i),
\end{aligned}
\right.
\end{equation}

\noindent where \( \hat{F}_i^{att} \), \( \hat{F}_i^{mlp} \), and \( \hat{H}_i \) are the projected representations of the original features.

\subsubsection{Adaptive Feature Aggregation.}

To compute feature weights, we leverage an adaptive weighting mechanism based on the final-layer hidden state \( H_n \). We use a Weight MLP to extract the weights from \( H_n \), and then apply them to all features through a weighted sum. The weighting process is defined as:

\begin{align}
w_i^{att},\; w_i^{mlp},\; w_i^H 
&= \sigma\big( \text{MLP}_{\text{weight}}(H_n) \big),\forall i &\in \{1,\ldots,n\}
\end{align}

\begin{equation}
F^{\text{agg}} = \sum_{i=1}^{n} \left( w_i^{att} \cdot \hat{F}_i^{att} + w_i^{mlp} \cdot \hat{F}_i^{mlp} + w_i^H \cdot \hat{H}_i \right),
\end{equation}

\noindent
where \( w_i^{att}, w_i^{mlp}, w_i^H \) are the learned adaptive weights for each type of feature, and \( F^{\text{agg}} \) is the aggregated feature representation used for classification; \( \sigma \) denotes the softmax function.

\subsection{Support-Query Attention Block}
Support and query samples are passed through the above blocks to extract their features and compute class prototypes following the PN method (Eq.~\eqref{eq1}). We denote the prototype of a batch's support set as \( s \) and the feature of a query sample as \( q \). Due to slight variations between the images in the support set and the query set (such as differences in background, lighting, or shooting angles), the prototypes generated from the support set may have a certain distance from the corresponding query set of the same class. This distribution bias leads to classification inaccuracy.

To mitigate this distribution bias, we propose the SQ Attention Block to adjust the position of the prototypes, making them closer to the distribution center of the corresponding query set, as shown in Figure \ref{fig:tsne}. Besides, before computing the correlation between prototype \( s \) and query \( q \), we apply a learnable projection \( \text{Proj}(\cdot) \) to \( q \) for adaptive, class-aware alignment, enhancing robustness under distribution shift. The formulation is as follows:

\begin{equation}
s^{att} = \alpha (s \cdot \text{Proj}(q)^T) \cdot q + (1-\alpha) \cdot s.
\label{eq13}
\end{equation}

Thus, the updated prototype \( s^{att} \) is obtained. Finally, the cosine similarity between each query sample and \( s^{att} \) is computed to produce the predicted labels.

\section{Experiments}
\subsection{Datasets}

Our work conducts experiments on four common FSL benchmarks, i.e., miniImageNet \cite{Vinyals2016}, tieredImageNet \cite{Ren2018}, CIFAR \cite{krizhevsky2009learning} and FC100 \cite{oreshkin2018tadam}, in which miniImageNet and tieredImageNet are derivatives of the ImageNet-1K dataset \cite{Russakovsky2015}. And we also conduct cross-domain few shot learning test on six datasets CUB \cite{Wah2011}, Stanford Cars \cite{krause20133d}, Places \cite{zhou2017places}, Plantae \cite{van2018inaturalist}, EuroSAT \cite{helber2019eurosat} and CropDiseases \cite{mohanty2016using}.
%-------------------------------------------------------------------------
\subsection{Setting}
\label{sec:5.2}

We utilize three backbone models: ViT-S/16 and ViT-B/16 pre-trained on the ImageNet-1K dataset, and ViT-B/16 pre-trained on the ImageNet-21K dataset. The $\xi$, $\zeta$, and $\alpha$ are tuned within the range $\{0.1, 1\}$. The hidden size of all bottleneck projections is set to 48, while the hidden size of the attention bottleneck projections for the query, key, and value in Active Block is set to 8. Following common practice~\cite{jie2023fact}, we adopt AdamW~\cite{loshchilov2017decoupled} as the optimizer and use a cosine learning rate scheduler. We set the batch size to 64 and optimized the model using a learning rate of 0.0001 for 5 epochs. For evaluation, we tested the model on 320 batches and reported the average performance along with the 95\% confidence interval. For data augmentation, we first resize the input image to $256 \times 256$ and then apply a center crop to $224 \times 224$. All experiments are conducted on a single NVIDIA V100 GPU.

%-------------------------------------------------------------------------
\subsection{Baseline}

State-of-the-art (SOTA) FSL methods are applied to the few-shot classification task to facilitate a thorough comparison, including FewTURE\cite{hiller2022rethinking}, MetaFormer-A \cite{yang2024one}, FewVS \cite{li2024fewvs}, SCAM-Net\cite{di2025brain}, SP\cite{chen2023semantic}, KTPP \cite{li2024knn}, SemFew \cite{zhang2024simple} as well as cross-domain SOTA methods like LDPNet\cite{zhou2023revisiting}, FLoR \cite{zou2024flatten}, StyleAdv \cite{fu2023styleadv}, AMT-FT \cite{yang2024mixture}. To further demonstrate the effectiveness of EfficientFSL under the PETL setting, we also compare against several commonly used PETL baselines Adapter \cite{chen2022vision}, AdaptFormer \cite{chen2022adaptformer}, and LoRA~\cite{hu2022lora}.

\subsection{Experimental Results and Analysis}
\subsubsection{Comparison with SOTA Methods.}

In Table \ref{tab:sota_comparison_all}, we evaluate our method on miniImageNet, tieredImageNet, CIFAR, and FC100 datasets, covering both the 5-way 1-shot and 5-shot settings. Our method demonstrates consistently superior performance across all settings and backbones, while using significantly fewer parameters.

Specifically, when using ViT-S, EfficientFSL surpasses the fully fine-tuned ViT-S and all recent works using the same pre-trained model. In particular, EfficientFSL only uses 1.25M parameters, which is substantially fewer than other ViT-based approaches, highlighting its efficiency. With ViT-B as the backbone and pre-trained on ImageNet-21K, EfficientFSL further improves the performance.

\begin{figure}[t]
  \centering
  \begin{subfigure}[b]{0.98\linewidth}
    \centering
    \includegraphics[width=\linewidth]{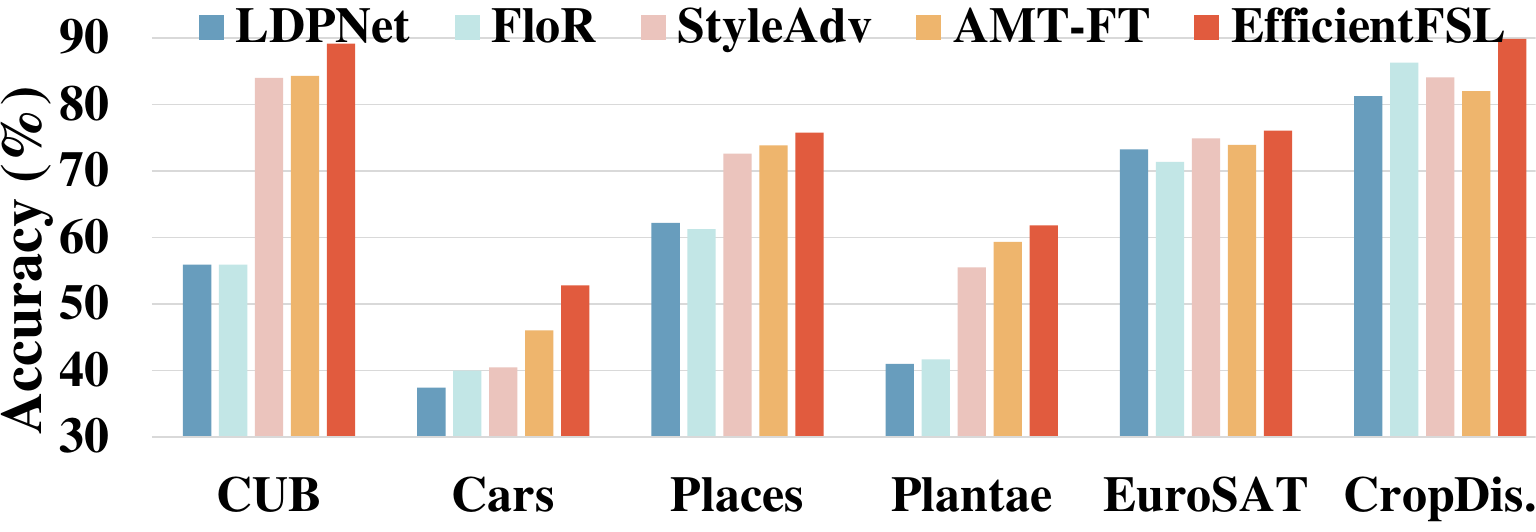}
    \caption{1-shot}
    \label{fig:cdfsl_1shot}
  \end{subfigure}
  \hfill
  \begin{subfigure}[b]{0.98\linewidth}
    \centering
    \includegraphics[width=\linewidth]{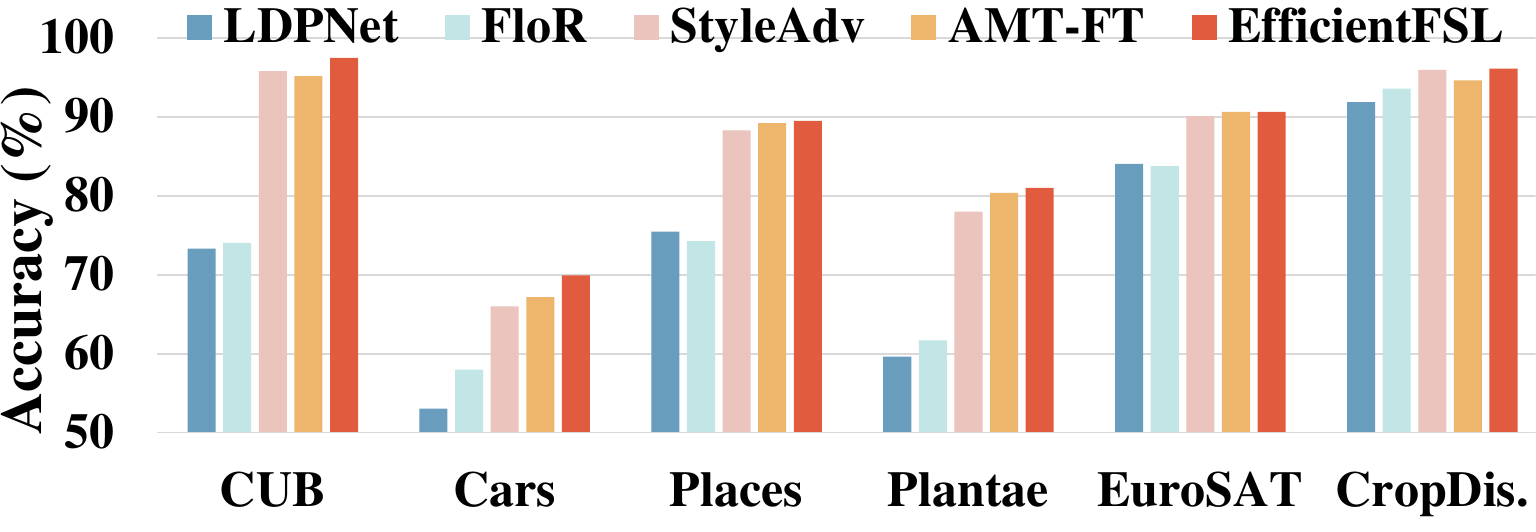}
    \caption{5-shot}
    \label{fig:cdfsl_5shot}
  \end{subfigure}
  \caption{Cross-domain Results under Different Shots.}
  \label{fig:cdfsl}
\end{figure}

To compare with SOTAs on cross-domain few-shot benchmarks, we evaluate the performance on six cross-domain few-shot benchmarks under the 5-way 1-shot and 5-way 5-shot scenario, using meta-learning on miniImageNet with a pre-trained ViT-S/16 model on ImageNet1K. As shown in Figure \ref{fig:cdfsl}, our approach outperforms existing SOTAs across all benchmarks, demonstrating its generalization capabilities in various domains.

Overall, these results underscore the effectiveness of EfficientFSL in various challenging FSL tasks, showing its superior generalization capabilities compared to recent SOTAs.

\subsubsection{Comparison with mainstream PETL approaches.}

To ensure a fair and rigorous comparison, we used the same backbone network on FC100 for all methods and carefully tuned their hyperparameters to maintain a comparable number of trainable parameters. 

As Table \ref{tab:petl} shows, EfficientFSL achieves significantly higher accuracy than the other compared methods on both ViT-S and ViT-B, demonstrating its superior performance. Concurrently, it also leads comprehensively in efficiency, achieving faster training and inference speeds and highly competitive memory usage.

\begin{table*}[t]
    
    \centering
    \fontsize{7.7}{11}\selectfont
    % 减少列间距，以适应更宽的表格
    \setlength{\tabcolsep}{2.3pt} 
    % 列数增加到 3 + 2*4 = 11 列
    \begin{tabular}{l c c cc cc cc cc}
        \toprule
        % 表头结构扩展，以包含所有四个数据集
        \multirow{2}{*}{Models} & \multirow{2}{*}{Backbone} & \multirow{2}{*}{Params} & \multicolumn{2}{c}{miniImageNet} & \multicolumn{2}{c}{tieredImageNet} & \multicolumn{2}{c}{CIFAR} & \multicolumn{2}{c}{FC100} \\  
        % \cmidrule 的范围也相应扩展
        \cmidrule(lr){4-5} \cmidrule(lr){6-7} \cmidrule(lr){8-9} \cmidrule(lr){10-11}
        & & & 5w-1s & 5w-5s & 5w-1s & 5w-5s & 5w-1s & 5w-5s & 5w-1s & 5w-5s \\  
        \midrule
        Full Fine-Tuning & ViT-S & 21.7M & 89.41 ± 0.88 & 95.59 ± 0.36 & 77.87 ± 1.35 & 89.13 ± 0.74 & 84.86 ± 1.11 & 92.30 ± 0.62 & 52.93 ± 1.29 & 66.70 ± 1.01 \\
        FewTURE (NeurIPS'22) & ViT-S & 21.7M & 68.02 ± 0.88 & 84.51 ± 0.53 & 72.96 ± 0.92 & 86.43 ± 0.67 & 72.80 ± 0.88 & 86.14 ± 0.64 & 46.20 ± 0.79 & 63.14 ± 0.73 \\
        MetaF. (ICML'24) & ViT-S & 24.5M & 84.78 ± 0.79 & 91.39 ± 0.42 & 88.38 ± 0.78 & 93.37 ± 0.45 & 88.34 ± 0.76 & 92.21 ± 0.59 & 58.04 ± 0.99 & 70.80 ± 0.76 \\
        FewVS (MM'24) & ViT-S & 21.7M & 86.80 ± 0.28 & 90.32 ± 0.22 & 87.87 ± 0.36 & 92.27 ± 0.26 & 85.63 ± 0.37 & 90.73 ± 0.33 & 61.01 ± 0.40 & 70.37 ± 0.39 \\
        SCAM-Net (arXiv'25) & ViT-S & 21.7M & 75.93 ± 0.45 & 89.75 ± 0.22 & 78.89 ± 0.52 & 88.67 ± 0.65 & 79.45 ± 0.53 & 90.90 ± 0.45 & 47.48 ± 0.32 & 65.85 ± 0.75 \\
        \rowcolor{lightblue}Ours(1K) & ViT-S & 1.25M & \textbf{97.40 ± 0.47} & \textbf{99.05 ± 0.14} & \textbf{89.72 ± 0.96} & \textbf{95.41 ± 0.48} & \textbf{88.82 ± 0.86} & \textbf{94.60 ± 0.42} & \textbf{69.94 ± 1.16} & \textbf{81.68 ± 0.95} \\
        \midrule
        % --- ViT-B Backbone Group ---
        Full Fine-Tuning & ViT-B & 85.8M & 90.51 ± 0.82 & 95.01 ± 0.37 & 80.17 ± 1.16 & 91.15 ± 0.58 & 85.83 ± 1.18 & 92.88 ± 0.59 & 52.86 ± 1.28 & 65.03 ± 0.99 \\
        SP (CVPR'23) & Visformer-T & 10.3M & 72.31 ± 0.40 & 83.42 ± 0.30 & 78.03 ± 0.46 & 88.55 ± 0.32 & 82.18 ± 0.40 & 88.24 ± 0.32 & 48.53 ± 0.38 & 61.55 ± 0.41 \\
        KTPP (MM’24) & Visformer-T & 11.1M & 76.71 ± 0.37 & 86.46 ± 0.27 & 80.80 ± 0.43 & 90.01 ± 0.29 & 83.63 ± 0.57 & 90.19 ± 0.30 & 51.59 ± 0.40 & 65.18 ± 0.40 \\
        SemFew (CVPR’24) & Swin-T & 88.0M & 78.94 ± 0.66 & 86.49 ± 0.50 & 82.37 ± 0.77 & 89.89 ± 0.52 & 84.34 ± 0.67 & 89.11 ± 0.54 & 54.27 ± 0.77 & 65.02 ± 0.72\\ 
        Ours(1K) & ViT-B & 2.48M & 97.57 ± 0.48 & 98.96 ± 0.14 & 87.63 ± 1.08 & 93.13 ± 0.62 & 85.25 ± 1.21 & 92.64 ± 0.54 & 72.60 ± 1.48 & 80.74 ± 1.04 \\
        \rowcolor{lightblue} Ours(21K) & ViT-B & 2.48M & \textbf{98.34 ± 0.30} & \textbf{99.12 ± 0.13} & \textbf{93.27 ± 0.75} & \textbf{96.78 ± 0.33} & \textbf{93.25 ± 0.69} & \textbf{97.28 ± 0.31} & \textbf{80.13 ± 1.26} & \textbf{88.81 ± 0.81} \\
        \bottomrule  
    \end{tabular}
    \caption{Comparison of our method with SOTAs on miniImageNet, tieredImageNet, CIFAR, and FC100 datasets. The “Params" column indicates the number of parameters (in millions). Ours (1K) refers to the full results with pre-trained ViT on ImageNet1K, while Ours (21K) uses ImageNet21K.}
    \label{tab:sota_comparison_all} % 新的标签
\end{table*}

\begin{table}[t]
\centering

\fontsize{7.7}{11}\selectfont
\renewcommand{\arraystretch}{0.9}
\setlength{\tabcolsep}{11.7pt}
% 删除了所有竖线 |
\begin{tabular}{l cccc} 
\toprule
{Method} & {Acc} $\uparrow$ & {TT} $\downarrow$ & {PM} $\downarrow$ & {IS} $\uparrow$ \\
\midrule
\midrule
\multicolumn{5}{c}{\textbf{Backbone: ViT-S}} \\
\midrule
Adapter & 62.80 & 26.68 & 0.50 & 380.46 \\
AdaptFormer & 64.13 & 38.22 & 0.57 & 274.53 \\
LoRA & 69.04 & 34.02 & 0.49 & 298.14 \\
\rowcolor{lightblue} EfficientFSL & \textbf{69.94} & \textbf{23.61} & \textbf{0.49} & \textbf{391.57} \\
\midrule
\midrule
\multicolumn{5}{c}{\textbf{Backbone: ViT-B}} \\
\midrule
Adapter & 67.48 & 74.19 & 1.09 & 134.21 \\
AdaptFormer & 68.83 & 76.87 & 1.19 & 129.98 \\
LoRA & 74.19 & 76.35 & 1.10 & 131.84 \\
\rowcolor{lightblue} EfficientFSL & \textbf{80.13} & \textbf{59.62} & \textbf{1.08} & \textbf{134.29} \\
\bottomrule
\end{tabular}
\caption{Comparison of different methods with ViT-S and ViT-B. Acc = Accuracy (\%), TT = Train Time (s/epoch), PM = Peak Memory (GB), IS = Inference Speed (images/s).}
\label{tab:petl}
\end{table}

\begin{table}[t]
    \centering
    \fontsize{7.7}{9}\selectfont
    \setlength{\tabcolsep}{7.15pt}
    \begin{tabular}{c c c ccl}
        \toprule
        Proj. & Att \& MLP & Combine &  Params (M) & 1-shot & 5-shot \\
        \midrule  
        \ding{55} & \ding{51} & \ding{51} & 1.58 & 51.15 & 68.95 \\
        \ding{51} & \ding{55} & \ding{51} & 1.05 & 72.15 & 87.81 \\
        \ding{51} & \ding{51} & \ding{55} & 2.37 & 75.55 & 88.60 \\
        \rowcolor{lightblue} \ding{51} & \ding{51} & \ding{51} & 2.48 & \textbf{80.13} & \textbf{88.81}  \\
        \bottomrule  
    \end{tabular}
    \caption{Impact of Removing Training Modules on Parameter Consumption and Performance. }
    \label{tab:re1}
\end{table}

\subsubsection{Impact of Removing Training Modules on Parameter Consumption and Performance.}

In Table \ref{tab:re1}, removing different components from the Active Block and Combine Block has a significant impact on both parameter consumption and performance.

When all modules are retained, the model achieves the best performance. Removing the projection layer (Proj.) reduces the parameter count but leads to a sharp drop in performance, indicating the essential role of the projection layer in ensuring high-quality feature representations. Removing the attention and MLP modules (Att \& MLP) in the Active Block reduces the parameter count by more than half, and performance drops. This shows that the attention mechanism and MLP modules are critical for capturing task-specific information and enhancing the model's performance. On the other hand, removing the Combine Block modules results in a smaller reduction in parameters, and the accuracy drop is relatively minor. 

These results highlight the importance of the projection layer and attention mechanism in the Active Block, as well as the crucial role of the Combine Block in efficient feature fusion, indicating its effectiveness in integrating multi-scale features.

\begin{table}[ht]

    \fontsize{7.7}{11}\selectfont
    \renewcommand{\arraystretch}{0.85}
    \setlength{\tabcolsep}{6.45pt}
    \centering
    \begin{tabular}{l|l|c c c}
    \toprule
    Module & Setting & Params (M) & 1-shot & 5-shot \\
    \midrule
    \multirow{3}{*}{Active Block} & w/o \( p_i \) & 2.44 & 39.36 & 60.99 \\
                            & w/o \( Att \) & 1.95 & 75.98 & 87.83 \\
                            & w/o \( MLP \) & 1.56 & 55.09 & 69.99 \\
    \midrule
    \multirow{3}{*}{Combine Block} & w/o \( F_i^{att} \) & 2.47 & 78.31 & \textbf{89.71} \\
                             & w/o \( F_i^{mlp} \) & 2.47 & 73.26 & 87.52 \\
                             & w/o \( H_i \) & 2.47 & 77.77 & 89.20 \\
    \midrule
    % \multirow{1}{*}{SQA Block} & w/o proj. & 2.47 & 78.31 & \textbf{89.71} \\
    % \midrule
    \rowcolor{lightblue} EfficientFSL & Complete Model & 2.48 & \textbf{80.13} & 88.81  \\
    \bottomrule  
    \end{tabular}
    \caption{Ablation Study on the Internal Components of Active Block and Combine Block.}
    \label{tab:re2}
\end{table}

\begin{table}[ht]
\centering
\fontsize{8}{11}\selectfont
\setlength{\tabcolsep}{2.21pt}
\begin{tabular}{l c cc cc cc cc}
\toprule
\multirow{2}{*}{Setting} & \multirow{2}{*}{\raisebox{-3ex}{\makecell{Params\\(M)}}} & \multicolumn{2}{c}{Mini} & \multicolumn{2}{c}{Tiered} & \multicolumn{2}{c}{CIFAR} & \multicolumn{2}{c}{FC100} \\
\cmidrule(lr){3-4} \cmidrule(lr){5-6} \cmidrule(lr){7-8} \cmidrule(lr){9-10}
                         &                         & 1shot   & 5shot    & 1shot    & 5shot     & 1shot    & 5shot     & 1shot   & 5shot    \\
\midrule
w/o SQA                  & 2.48                    & 95.95   & 98.97    & 89.78    & 96.61     & 90.15    & 96.99     & 75.42   & 87.82    \\
w/o proj                 & 2.48                    & 98.34   & 99.12    & 93.27    & 96.78     & 93.25    & 97.28     & 80.13   & 88.81    \\
\rowcolor{lightblue} Ours                   & 2.55& \textbf{98.49} & \textbf{99.23} & \textbf{93.85} & \textbf{96.89} & \textbf{93.34} & \textbf{97.28} & \textbf{80.20} & \textbf{89.53} \\
\bottomrule
\end{tabular}
\caption{Ablation Analysis of the SQ Attention Block.}
\label{tab:SQA}
\end{table}

\begin{figure}[h]
  \centering
  \includegraphics[width=0.7\linewidth]{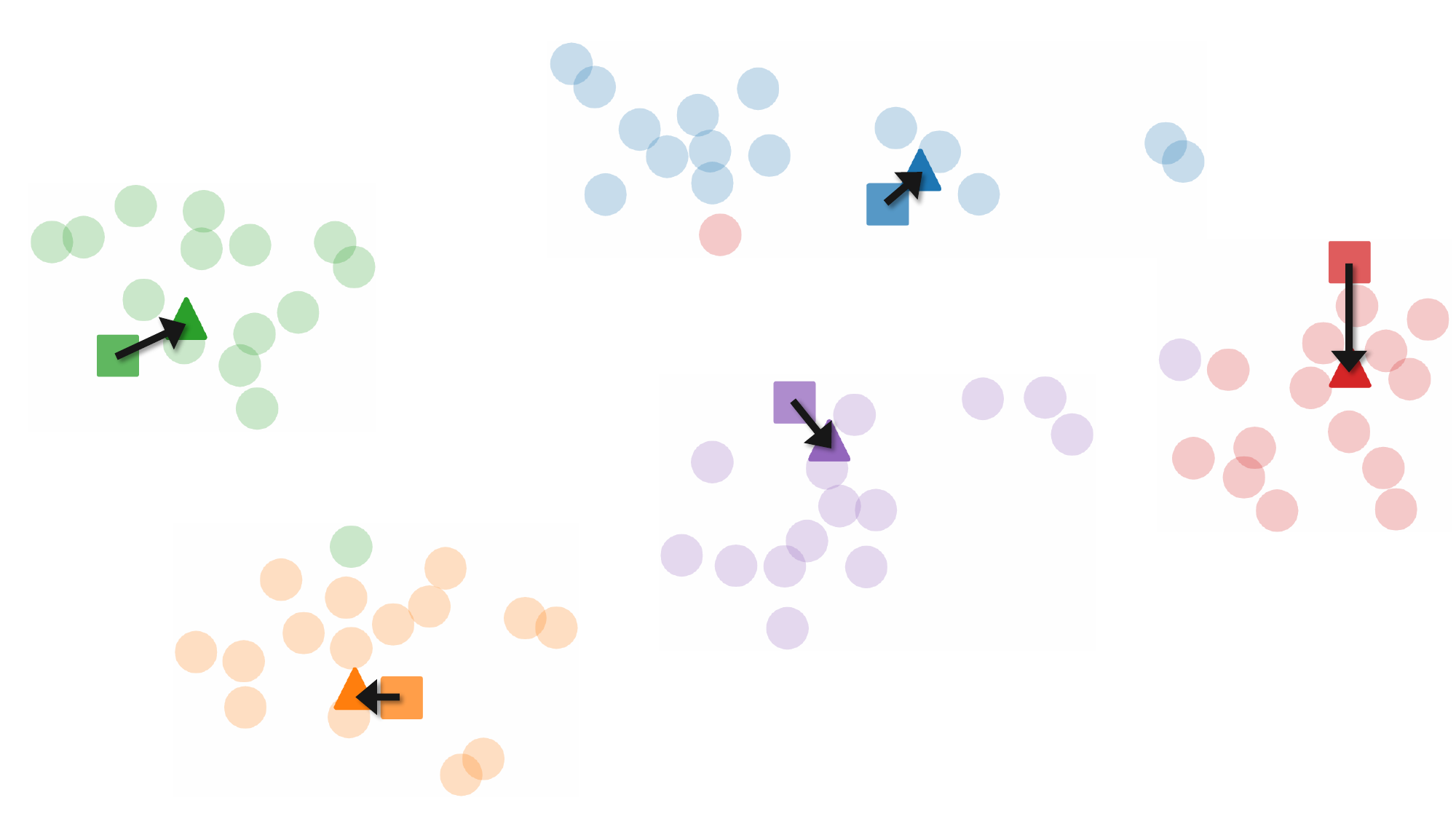}
  \caption{t-SNE visualization of a 5-way 1-shot 15-query task on the FC100 dataset. Squares represent prototypes computed by simply averaging support features (\( s^{c} \)); triangles represent prototypes computed by SQ Attention Block (\( s^{att} \)); circles denote query samples.}
  \label{fig:tsne}
\end{figure}

\subsubsection{Ablation Study on Active and Combine Block Components.}

Table~\ref{tab:re2} analyzes the impact of independently removing key components from the Active and Combine Blocks on parameter count and performance.

For the Active Block, removing $p_i$ causes a significant performance drop with negligible parameter reduction, confirming its critical role in guiding learning. Ablating $Att$ slightly reduces accuracy alongside parameter savings, while removing $MLP$ leads to a larger performance decrease, underscoring its importance in refining features. Overall, all three modules are essential for efficient feature learning.

For the Combine Block, ablating $F_i^{att}$ has limited effect, mainly on 1-shot tasks; removing $F_i^{mlp}$ yields a larger drop, highlighting its role in decision adjustment; removing $H_i$ causes the largest performance loss, showing its key contribution to knowledge retention. While parameter reduction is minimal, multi-scale outputs remain important for performance. These highlight the contribution of each component and provide guidance for future model optimization.

\subsubsection{Impact of SQ Attention Block.}

As shown in Table \ref{tab:SQA}, removing the SQ Attention Block leads to a decrease in accuracy, demonstrating the substantial impact of the SQ Attention Block on performance. Moreover, removing projection on \( q \) results in a slight yet consistent performance drop. This confirms that adjusting \( q \) with a class-aware projection layer prior to the correlation calculation with \( s \) alleviates the adverse effects of inter-class cluster overlap.

Figure \ref{fig:tsne} shows the t-SNE visualization of one batch of test samples from the experiment. The figure clearly demonstrates that SQ Attention Block, through the attention mechanism, brings the prototypes closer to the distribution center of the corresponding class's query samples. This highlights that the SQ Attention Block significantly reduces the distribution shift between the support set and the query set.

\subsubsection{Impact of Different Feature Aggregation Methods on Parameter Size and Few-Shot Performance.}

Table~\ref{tab:combine} compares three feature aggregation strategies in the Combine Block and their impact on parameter size and few-shot performance. Simple averaging treats all layers equally, while fixed weights introduce input-invariant learnable parameters with slight improvements. In contrast, conditional weights dynamically generate aggregation weights based on the final-layer output, achieving the best performance with minimal parameter increase (2.48M). 

These results show that adaptive, condition-aware weighting significantly improves generalization in FSL, especially under low-data settings. With minimal parameter overhead, it provides an efficient yet effective alternative to simpler aggregation strategies.

\begin{table}[t]

    \centering
    \fontsize{8.5}{11}\selectfont
    \setlength{\tabcolsep}{4.51pt}
    \begin{tabular}{c c c c}
        \toprule
        Combine Block & Params (M) & 1shot & 5shot \\
        \midrule  
        Simple Average & 2.45 & 69.58 ± 1.22 & 85.38 ± 0.83 \\
        Fixed Weights & 2.45 & 69.97 ± 1.22 & 85.49 ± 0.84 \\
        \rowcolor{lightblue} Conditional Weights & 2.48 & \textbf{80.13 ± 1.26} & \textbf{88.81 ± 0.81} \\
        \bottomrule  
    \end{tabular}
    \caption{Ablation of Different Feature Fusion Methods.}
    \label{tab:combine}
\end{table}

\begin{figure}[t] % 单栏展示
  \centering

  % 第一行
  \begin{subfigure}[b]{0.49\linewidth}
    \centering
    \includegraphics[width=\linewidth]{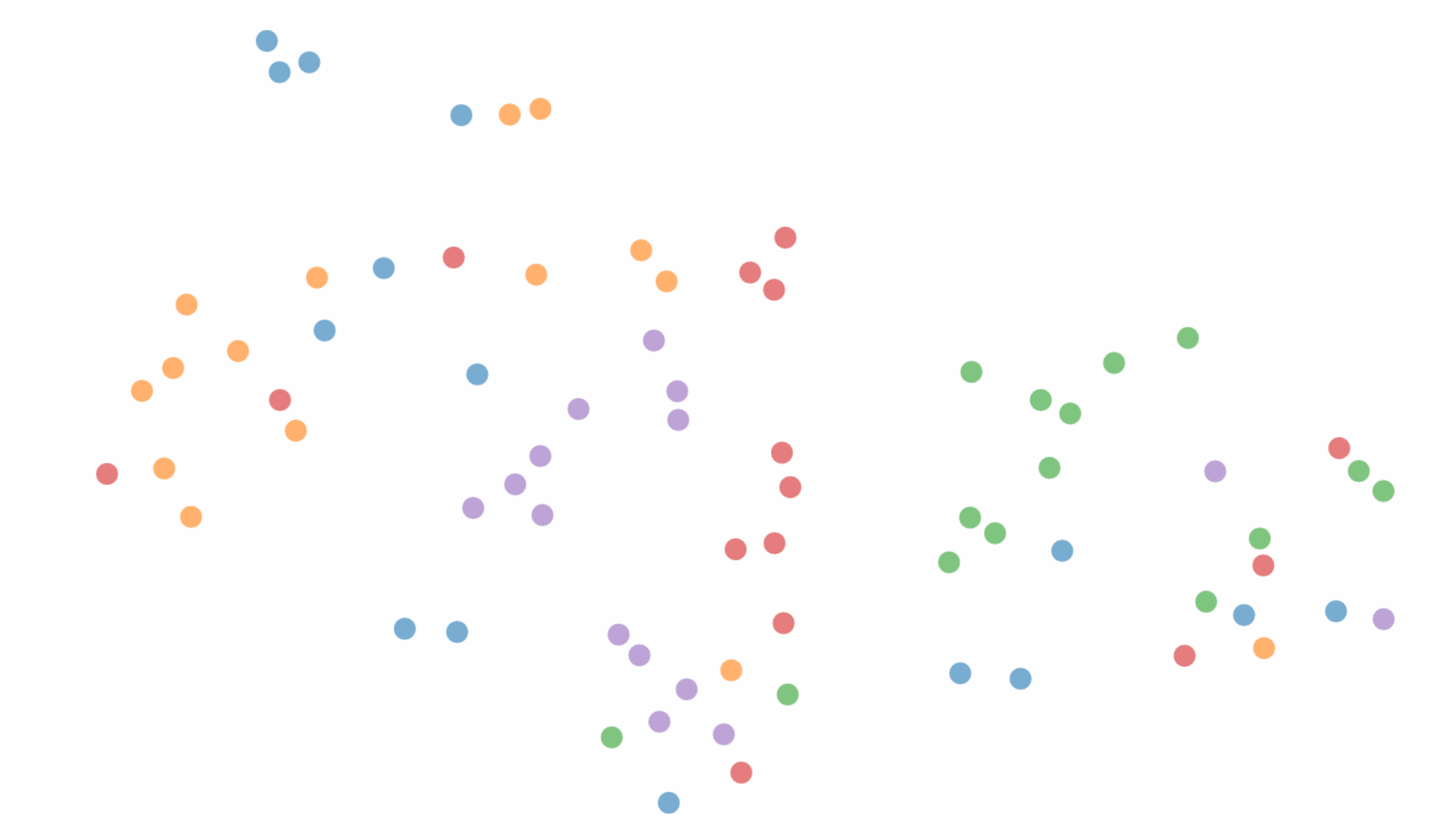}
    \caption{w/o Fine-tuning}
    \label{fig:tsne_a}
  \end{subfigure}
  \hfill
  \begin{subfigure}[b]{0.49\linewidth}
    \centering
    \includegraphics[width=\linewidth]{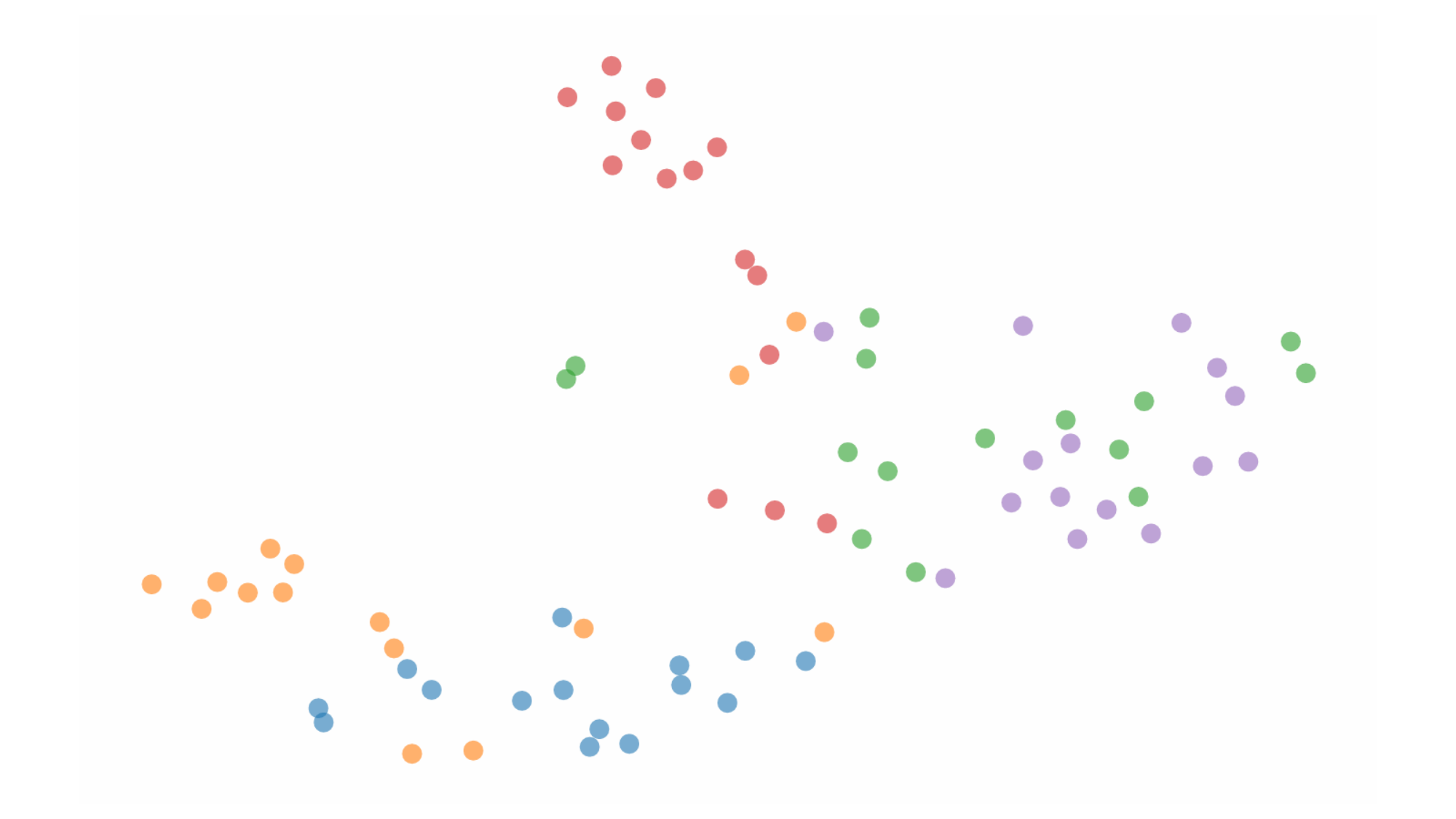}
    \caption{w/ Prompt $P_i$}
    \label{fig:tsne_b}
  \end{subfigure}

  \vspace{0.5em} % 调整行距

  % 第二行
  \begin{subfigure}[b]{0.49\linewidth}
    \centering
    \includegraphics[width=\linewidth]{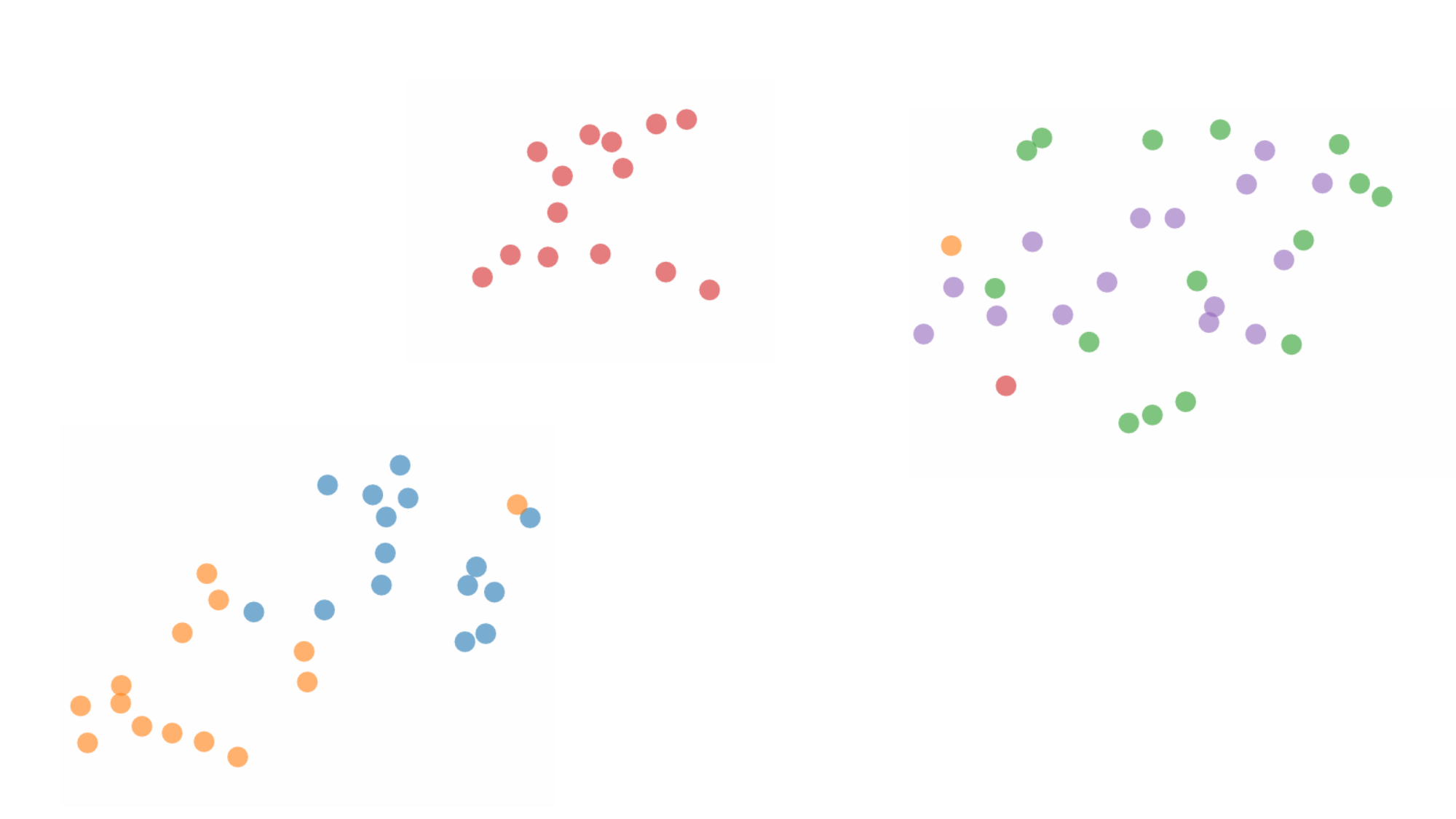}
    \caption{w/ $P_i$ and $\text{Proj}(\cdot)$}
    \label{fig:tsne_c}
  \end{subfigure}
  \hfill
  \begin{subfigure}[b]{0.49\linewidth}
    \centering
    \includegraphics[width=\linewidth]{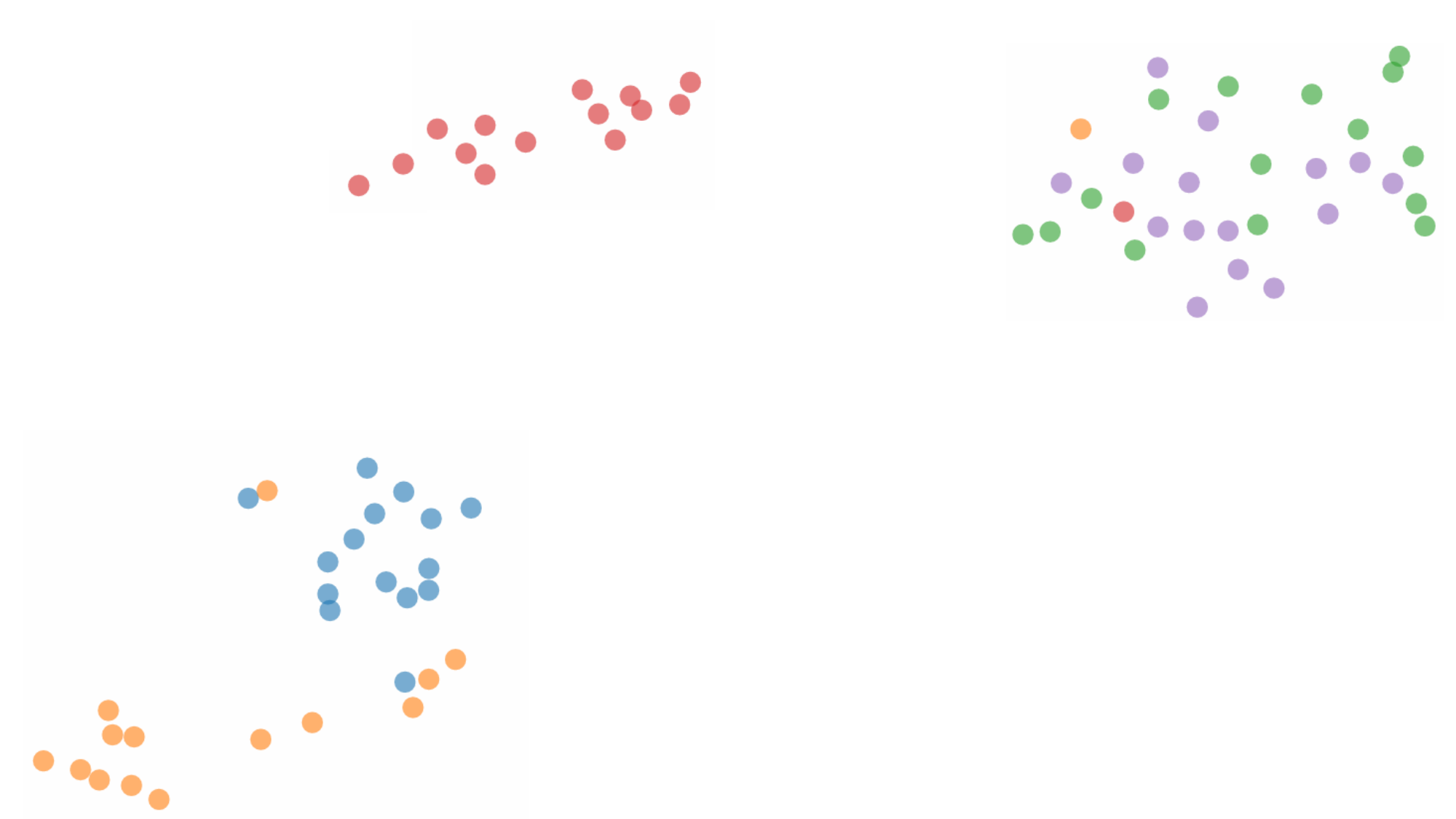}
    \caption{w/ Active Block}
    \label{fig:tsne_d}
  \end{subfigure}

  \vspace{0.5em}

  % 第三行
  \begin{subfigure}[b]{0.49\linewidth}
    \centering
    \includegraphics[width=\linewidth]{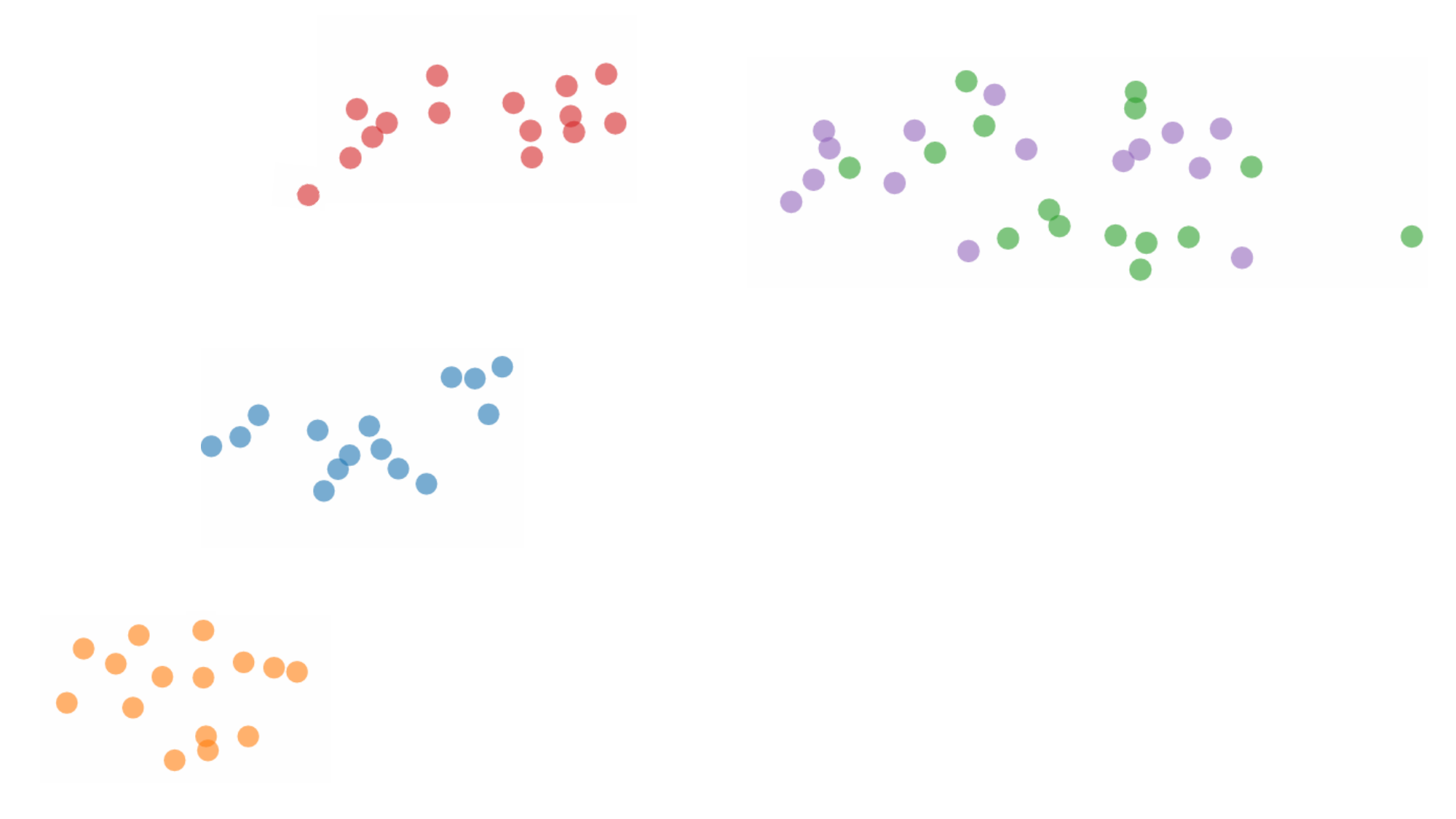}
    \caption{w/ Active \& Combine Block}
    \label{fig:tsne_e}
  \end{subfigure}
  \hfill
  \begin{subfigure}[b]{0.49\linewidth}
    \centering
    \includegraphics[width=\linewidth]{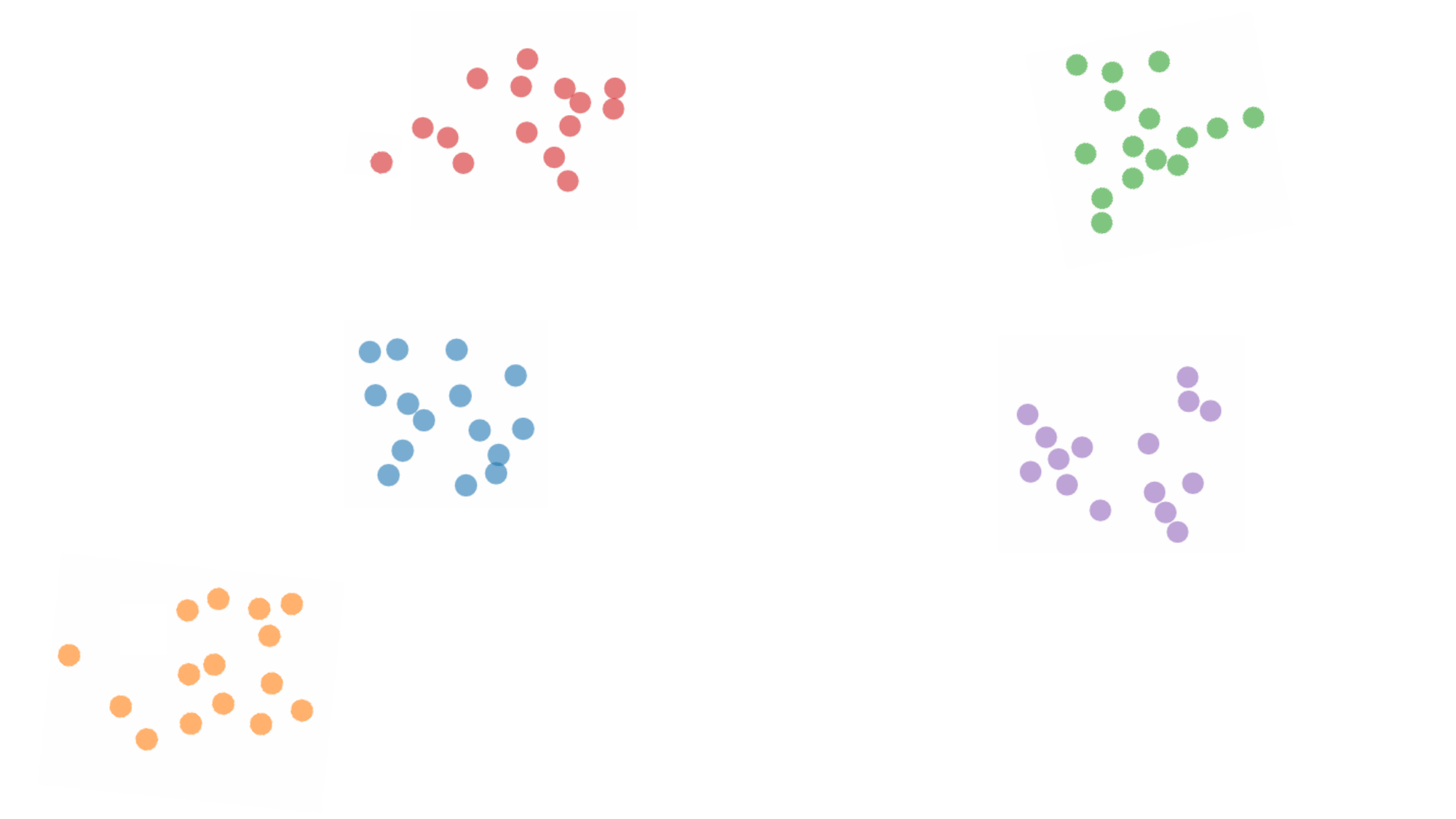}
    \caption{EfficientFSL (w/ SQAttn)}
    \label{fig:tsne_f}
  \end{subfigure}

  \caption{t-SNE visualizations of test query samples under six settings. (b)-(f) incrementally add components to the previous setting.}
  \label{fig:tsne_all}
\end{figure}

\subsubsection{From Chaos to Order: Evolution of the Representation Space through Module Integration.}

Figure~\ref{fig:tsne_all} shows how the feature space is progressively refined as different modules are incrementally introduced. Starting from the pre-trained ViT-B/16 without fine-tuning (Figure~\ref{fig:tsne_a}), query features are highly entangled with vague class boundaries. Adding trainable prompts ($P_i$) (Figure~\ref{fig:tsne_b}) induces coarse clustering, showing initial task adaptation but with noticeable inter-class overlap. Introducing the projection layer ($\text{Proj}(\cdot)$) (Figure~\ref{fig:tsne_c}) further improves separability and cluster compactness. With the integration of self-attention and MLP into an Active Block (Figure~\ref{fig:tsne_d}), clusters become tighter and class boundaries clearer. Adding the Combine Block (Figure~\ref{fig:tsne_e}) enhances intra-class cohesion and inter-class separation. Finally, the SQ Attention Block dynamically adjusts prototypes, yielding the full EfficientFSL framework (Figure~\ref{fig:tsne_f}), where features form clear, well-separated clusters. These results validate the effectiveness of our modular design in improving representation quality for FSL.

\section{Conclusion}

To bridge the gap between powerful pre-trained representations and the limited resources in FSL, we propose \textbf{EfficientFSL}, a parameter-efficient fine-tuning framework for few-shot classification. By introducing the Forward Block, Combine Block, and SQ Attention Block, EfficientFSL enables task-specific adaptation with minimal trainable parameters. 
Extensive experiments on both in-domain and cross-domain benchmarks show that EfficientFSL consistently outperforms existing methods, exhibiting strong performance and generalization capabilities. In particular, the SQ Attention Block effectively mitigates support-query distribution shift, thereby enhancing the effectiveness of prototype-based classification. Overall, EfficientFSL strikes a strong balance between efficiency, scalability, and performance, offering an effective solution for FSL applications.

\section{Acknowledgments}
This work is supported by the National Natural Science Foundation of China (Grant Number: 62203118) and the Fundamental Research Funds for the Central Universities (Grant Number: 2025SMECP012).

\bibliography{aaai2026}

@String{Computer = "{IEEE} Computer" }

@String{Chelsea = "Chelsea" }

@String{Springer = "Springer-Verlag" }

@BOOK{test,
   author = "Donald E. Knuth",
   title = "Seminumerical Algorithms",
   volume = 2,
   series = "The Art of Computer Programming",
   publisher = "Addison-Wesley",
   address = "Reading, MA",
   edition = "2nd",
   month = "10~" # jan,
   year = "1981",
}

@inproceedings{yang2024one,
  title={One meta-tuned transformer is what you need for few-shot learning},
  author={Yang, Xu and Yao, Huaxiu and Wei, Ying},
  booktitle={Forty-first International Conference on Machine Learning},
  year={2024}
}

@inproceedings{li2024fewvs,
  title={FewVS: A Vision-Semantics Integration Framework for Few-Shot Image Classification},
  author={Li, Zhuoling and Wang, Yong and Li, Kaitong},
  booktitle={Proceedings of the 32nd ACM International Conference on Multimedia},
  pages={1341--1350},
  year={2024}
}

@inproceedings{zhang2024simple,
  title={Simple semantic-aided few-shot learning},
  author={Zhang, Hai and Xu, Junzhe and Jiang, Shanlin and He, Zhenan},
  booktitle={Proceedings of the IEEE/CVF Conference on Computer Vision and Pattern Recognition},
  pages={28588--28597},
  year={2024}
}

@inproceedings{li2024knn,
  title={KNN Transformer with Pyramid Prompts for Few-Shot Learning},
  author={Li, Wenhao and Wang, Qiangchang and Zhao, Peng and Yin, Yilong},
  booktitle={Proceedings of the 32nd ACM International Conference on Multimedia},
  pages={1082--1091},
  year={2024}
}

@inproceedings{fu2023styleadv,
    title={StyleAdv: Meta Style Adversarial Training for Cross-Domain Few-Shot
           Learning},
    author={Fu, Yuqian and Xie, Yu and Fu, Yanwei and Jiang, Yu-Gang},
    booktitle={Proceedings of the IEEE/CVF Conference on Computer Vision and
               Pattern Recognition},
    pages={24575--24584},
    year={2023}
}

@inproceedings{zou2024flatten,
  title={Flatten long-range loss landscapes for cross-domain few-shot learning},
  author={Zou, Yixiong and Liu, Yicong and Hu, Yiman and Li, Yuhua and Li, Ruixuan},
  booktitle={Proceedings of the IEEE/CVF Conference on Computer Vision and Pattern Recognition},
  pages={23575--23584},
  year={2024}
}

@inproceedings{yang2024mixture,
  title={Mixture of Adversarial LoRAs: Boosting Robust Generalization in Meta-Tuning},
  author={Yang, Xu and Liu, Chen and Wei, Ying},
  booktitle={The Thirty-eighth Annual Conference on Neural Information Processing Systems},
  year={2024}
}

@article{zhang2022deepemd,
  title={Deepemd: Differentiable earth mover's distance for few-shot learning},
  author={Zhang, Chi and Cai, Yujun and Lin, Guosheng and Shen, Chunhua},
  journal={IEEE Transactions on Pattern Analysis and Machine Intelligence},
  volume={45},
  number={5},
  pages={5632--5648},
  year={2022},
  publisher={IEEE}
}

@inproceedings{afrasiyabi2022matching,
  title={Matching feature sets for few-shot image classification},
  author={Afrasiyabi, Arman and Larochelle, Hugo and Lalonde, Jean-Fran{\c{c}}ois and Gagn{\'e}, Christian},
  booktitle={Proceedings of the IEEE/CVF conference on computer vision and pattern recognition},
  pages={9014--9024},
  year={2022}
}

@inproceedings{finn2017model,
  title={Model-agnostic meta-learning for fast adaptation of deep networks},
  author={Finn, Chelsea and Abbeel, Pieter and Levine, Sergey},
  booktitle={International conference on machine learning},
  pages={1126--1135},
  year={2017},
  organization={PMLR}
}

@inproceedings{munkhdalai2018rapid,
  title={Rapid adaptation with conditionally shifted neurons},
  author={Munkhdalai, Tsendsuren and Yuan, Xingdi and Mehri, Soroush and Trischler, Adam},
  booktitle={International conference on machine learning},
  pages={3664--3673},
  year={2018},
  organization={PMLR}
}

@article{antoniou2018train,
  title={How to train your MAML},
  author={Antoniou, Antreas and Edwards, Harrison and Storkey, Amos},
  journal={arXiv preprint arXiv:1810.09502},
  year={2018}
}

@article{dosovitskiy2020image,
  title={An image is worth 16x16 words: Transformers for image recognition at scale},
  author={Dosovitskiy, Alexey and Beyer, Lucas and Kolesnikov, Alexander and Weissenborn, Dirk and Zhai, Xiaohua and Unterthiner, Thomas and Dehghani, Mostafa and Minderer, Matthias and Heigold, Georg and Gelly, Sylvain and others},
  journal={arXiv preprint arXiv:2010.11929},
  year={2020}
}

@inproceedings{zhai2022scaling,
  title={Scaling vision transformers},
  author={Zhai, Xiaohua and Kolesnikov, Alexander and Houlsby, Neil and Beyer, Lucas},
  booktitle={Proceedings of the IEEE/CVF conference on computer vision and pattern recognition},
  pages={12104--12113},
  year={2022}
}

@article{hu2022lora,
  title={Lora: Low-rank adaptation of large language models.},
  author={Hu, Edward J and Shen, Yelong and Wallis, Phillip and Allen-Zhu, Zeyuan and Li, Yuanzhi and Wang, Shean and Wang, Lu and Chen, Weizhu and others},
  journal={ICLR},
  volume={1},
  number={2},
  pages={3},
  year={2022}
}

@inproceedings{jia2022visual,
  title={Visual prompt tuning},
  author={Jia, Menglin and Tang, Luming and Chen, Bor-Chun and Cardie, Claire and Belongie, Serge and Hariharan, Bharath and Lim, Ser-Nam},
  booktitle={European conference on computer vision},
  pages={709--727},
  year={2022},
  organization={Springer}
}

@inproceedings{jie2023fact,
  title={Fact: Factor-tuning for lightweight adaptation on vision transformer},
  author={Jie, Shibo and Deng, Zhi-Hong},
  booktitle={Proceedings of the AAAI conference on artificial intelligence},
  volume={37},
  number={1},
  pages={1060--1068},
  year={2023}
}

@inproceedings{jie2023revisiting,
  title={Revisiting the parameter efficiency of adapters from the perspective of precision redundancy},
  author={Jie, Shibo and Wang, Haoqing and Deng, Zhi-Hong},
  booktitle={Proceedings of the IEEE/CVF International Conference on Computer Vision},
  pages={17217--17226},
  year={2023}
}

@inproceedings{Ravi2017,
  author    = {Sachin Ravi and Hugo Larochelle},
  title     = {Optimization as a Model for Few-Shot Learning},
  booktitle = {International Conference on Learning Representations (ICLR)},
  year      = {2017}
}

@inproceedings{Rusu2019,
  author    = {Andrei A. Rusu and Dushyant Rao and Jakub Sygnowski and Oriol Vinyals and Razvan Pascanu and Simon Osindero and Raia Hadsell},
  title     = {Meta-Learning with Latent Embedding Optimization},
  booktitle = {International Conference on Learning Representations (ICLR)},
  year      = {2019}
}

@inproceedings{Baik2020,
  author    = {Sungyong Baik and Seokil Hong and Kyoung Mu Lee},
  title     = {Learning to Forget for Meta-Learning},
  booktitle = {IEEE Conference on Computer Vision and Pattern Recognition (CVPR)},
  pages     = {2376--2384},
  year      = {2020}
}

@inproceedings{Wang2018,
  author    = {Yu-Xiong Wang and Ross B. Girshick and Martial Hebert and Bharath Hariharan},
  title     = {Low-shot Learning from Imaginary Data},
  booktitle = {IEEE Conference on Computer Vision and Pattern Recognition (CVPR)},
  pages     = {7278--7286},
  year      = {2018}
}

@inproceedings{Zhang2019,
  author    = {Hongguang Zhang and Jing Zhang and Piotr Koniusz},
  title     = {Few-shot Learning via Saliency-guided Hallucination of Samples},
  booktitle = {IEEE Conference on Computer Vision and Pattern Recognition (CVPR)},
  pages     = {2770--2779},
  year      = {2019}
}

@inproceedings{Wang2020,
  author    = {Haoran Wang and Ying Zhang and Zhong Ji and Yanwei Pang and Lin Ma},
  title     = {Consensus-aware Visual-Semantic Embedding for Image-Text Matching},
  booktitle = {European Conference on Computer Vision (ECCV)},
  volume    = {12369},
  pages     = {18--34},
  year      = {2020}
}

@inproceedings{Snell2017,
  author    = {Jake Snell and Kevin Swersky and Richard Zemel},
  title     = {Prototypical Networks for Few-Shot Learning},
  booktitle = {Advances in Neural Information Processing Systems (NeurIPS)},
  pages     = {4077--4087},
  year      = {2017}
}

@inproceedings{Vinyals2016,
  author    = {Oriol Vinyals and Charles Blundell and Timothy Lillicrap and Daan Wierstra},
  title     = {Matching Networks for One Shot Learning},
  booktitle = {Advances in Neural Information Processing Systems (NeurIPS)},
  pages     = {3630--3638},
  year      = {2016}
}

@inproceedings{Koch2015,
  author    = {Gregory Koch and Richard Zemel and Ruslan Salakhutdinov},
  title     = {Siamese neural networks for oneshot image recognition},
  booktitle = {ICML deep learning workshop},
  volume    = {2},
  year      = {2015},
  address   = {Lille}
}

@inproceedings{Chen2020,
  author    = {Jiaxin Chen and Li-Ming Zhan and Xiao-Ming Wu and Fu-lai Chung},
  title     = {Variational metric scaling for metric-based meta-learning},
  booktitle = {Proceedings of the AAAI Conference on Artificial Intelligence},
  volume    = {34},
  pages     = {3478--3485},
  year      = {2020}
}

@article{Fu2022,
  author    = {L. Fu and others},
  title     = {Learning Robust Discriminant Subspace Based on Joint L2,p and L2,s-Norm Distance Metrics},
  journal   = {IEEE Transactions on Neural Networks and Learning Systems},
  volume    = {33},
  number    = {1},
  pages     = {130--144},
  year      = {2022},
  month     = {Jan}
}

@article{Tang2022,
  author    = {H. Tang and C. Yuan and Z. Li and J. Tang},
  title     = {Learning Attention-Guided Pyramidal Features for Few-Shot Fine-Grained Recognition},
  journal   = {Pattern Recognition},
  volume    = {130},
  pages     = {108792},
  year      = {2022},
  month     = {Oct}
}

@inproceedings{Sung2018,
  author    = {F. Sung and Y. Yang and L. Zhang and T. Xiang and P. H. S. Torr and T. M. Hospedales},
  title     = {Learning to Compare: Relation Network for Few-Shot Learning},
  booktitle = {IEEE/CVF Conference on Computer Vision and Pattern Recognition (CVPR)},
  pages     = {1199--1208},
  year      = {2018},
  month     = {Jun}
}

@inproceedings{Liu2020,
  author    = {B. Liu and others},
  title     = {Negative Margin Matters: Understanding Margin in Few-Shot Classification},
  booktitle = {European Conference on Computer Vision (ECCV)},
  pages     = {438--455},
  year      = {2020}
}

@inproceedings{Ren2018,
  author = {Ren, M. and Triantafillou, E. and Ravi, S. and Snel1, J. and Swersky, K. and Tenenbaum, J. B. and Larochelle, H. and Zemel, R. S.},
  title = {Meta-learning for semi-supervised few-shot classification},
  booktitle = {International Conference on Learning Representations (ICLR)},
  year = {2018}
}

@misc{Wah2011,
  author = {Wah, C. and Branson, S. and Welinder, P. and Perona, P. and Belongie, S.},
  title = {The Caltech-UCSD Birds-200-2011 dataset},
  year = {2011},
}

@article{Russakovsky2015,
  author = {Russakovsky, O. and et al.},
  title = {ImageNet Large Scale Visual Recognition Challenge},
  journal = {International Journal of Computer Vision},
  volume = {115},
  number = {3},
  pages = {211--252},
  year = {2015},
  month = {December}
}

@article{zhou2017places,
  title={Places: A 10 million image database for scene recognition},
  author={Zhou, Bolei and Lapedriza, Agata and Khosla, Aditya and Oliva, Aude and Torralba, Antonio},
  journal={IEEE transactions on pattern analysis and machine intelligence},
  volume={40},
  number={6},
  pages={1452--1464},
  year={2017},
  publisher={IEEE}
}

@inproceedings{van2018inaturalist,
  title={The inaturalist species classification and detection dataset},
  author={Van Horn, Grant and Mac Aodha, Oisin and Song, Yang and Cui, Yin and Sun, Chen and Shepard, Alex and Adam, Hartwig and Perona, Pietro and Belongie, Serge},
  booktitle={Proceedings of the IEEE conference on computer vision and pattern recognition},
  pages={8769--8778},
  year={2018}
}

@article{krizhevsky2009learning,
  title={Learning multiple layers of features from tiny images},
  author={Krizhevsky, Alex and Hinton, Geoffrey and others},
  year={2009},
  publisher={Toronto, ON, Canada}
}

@article{oreshkin2018tadam,
  title={Tadam: Task dependent adaptive metric for improved few-shot learning},
  author={Oreshkin, Boris and Rodr{\'\i}guez L{\'o}pez, Pau and Lacoste, Alexandre},
  journal={Advances in neural information processing systems},
  volume={31},
  year={2018}
}

@inproceedings{krause20133d,
  title={3d object representations for fine-grained categorization},
  author={Krause, Jonathan and Stark, Michael and Deng, Jia and Fei-Fei, Li},
  booktitle={Proceedings of the IEEE international conference on computer vision workshops},
  pages={554--561},
  year={2013}
}

@article{loshchilov2017decoupled,
  title={Decoupled weight decay regularization},
  author={Loshchilov, Ilya and Hutter, Frank},
  journal={arXiv preprint arXiv:1711.05101},
  year={2017}
}

@article{li2021prefix,
  title={Prefix-tuning: Optimizing continuous prompts for generation},
  author={Li, Xiang Lisa and Liang, Percy},
  journal={arXiv preprint arXiv:2101.00190},
  year={2021}
}

@inproceedings{houlsby2019parameter,
  title={Parameter-efficient transfer learning for NLP},
  author={Houlsby, Neil and Giurgiu, Andrei and Jastrzebski, Stanislaw and Morrone, Bruna and De Laroussilhe, Quentin and Gesmundo, Andrea and Attariyan, Mona and Gelly, Sylvain},
  booktitle={International conference on machine learning},
  pages={2790--2799},
  year={2019},
  organization={PMLR}
}

@article{chen2022adaptformer,
  title={Adaptformer: Adapting vision transformers for scalable visual recognition},
  author={Chen, Shoufa and Ge, Chongjian and Tong, Zhan and Wang, Jiangliu and Song, Yibing and Wang, Jue and Luo, Ping},
  journal={Advances in Neural Information Processing Systems},
  volume={35},
  pages={16664--16678},
  year={2022}
}

@article{he2021towards,
  title={Towards a unified view of parameter-efficient transfer learning},
  author={He, Junxian and Zhou, Chunting and Ma, Xuezhe and Berg-Kirkpatrick, Taylor and Neubig, Graham},
  journal={arXiv preprint arXiv:2110.04366},
  year={2021}
}

@article{valipour2022dylora,
  title={Dylora: Parameter efficient tuning of pre-trained models using dynamic search-free low-rank adaptation},
  author={Valipour, Mojtaba and Rezagholizadeh, Mehdi and Kobyzev, Ivan and Ghodsi, Ali},
  journal={arXiv preprint arXiv:2210.07558},
  year={2022}
}

@article{zhang2023adalora,
  title={Adalora: Adaptive budget allocation for parameter-efficient fine-tuning},
  author={Zhang, Qingru and Chen, Minshuo and Bukharin, Alexander and Karampatziakis, Nikos and He, Pengcheng and Cheng, Yu and Chen, Weizhu and Zhao, Tuo},
  journal={arXiv preprint arXiv:2303.10512},
  year={2023}
}

@article{liu2022few,
  title={Few-shot parameter-efficient fine-tuning is better and cheaper than in-context learning},
  author={Liu, Haokun and Tam, Derek and Muqeeth, Mohammed and Mohta, Jay and Huang, Tenghao and Bansal, Mohit and Raffel, Colin A},
  journal={Advances in Neural Information Processing Systems},
  volume={35},
  pages={1950--1965},
  year={2022}
}

@article{helber2019eurosat,
  title     = {{EuroSAT}: A novel dataset and deep learning benchmark for land use and land cover classification},
  author    = {Helber, Patrick and Bischke, Benjamin and Dengel, Andreas and Borth, Damian},
  journal   = {{IEEE Journal of Selected Topics in Applied Earth Observations and Remote Sensing}},
  volume    = {12},
  number    = {7},
  pages     = {2217--2226},
  year      = {2019},
  publisher = {IEEE}
}

@article{mohanty2016using,
  title     = {Using deep learning for image-based plant disease detection},
  author    = {Mohanty, Sharada P and Hughes, David P and Salath{\'e}, Marcel},
  journal   = {{Frontiers in Plant Science}},
  volume    = {7},
  pages     = {215232},
  year      = {2016},
  publisher = {Frontiers}
}

@inproceedings{zhou2023revisiting,
  title={Revisiting prototypical network for cross domain few-shot learning},
  author={Zhou, Fei and Wang, Peng and Zhang, Lei and Wei, Wei and Zhang, Yanning},
  booktitle={Proceedings of the IEEE/CVF conference on computer vision and pattern recognition},
  pages={20061--20070},
  year={2023}
}

@article{hiller2022rethinking,
  title={Rethinking generalization in few-shot classification},
  author={Hiller, Markus and Ma, Rongkai and Harandi, Mehrtash and Drummond, Tom},
  journal={Advances in neural information processing systems},
  volume={35},
  pages={3582--3595},
  year={2022}
}

@article{di2025brain,
  title={Brain Inspired Adaptive Memory Dual-Net for Few-Shot Image Classification},
  author={Di, Kexin and Li, Xiuxing and Han, Yuyang and Li, Ziyu and Li, Qing and Wu, Xia},
  journal={arXiv preprint arXiv:2503.07396},
  year={2025}
}

@article{chen2023semantic,
  title={Semantic prompt for few-shot image recognition},
  author={Chen, Wentao and Si, Chenyang and Zhang, Zhang and Wang, Liang and Wang, Zilei and Tan, Tieniu},
  journal={arXiv preprint arXiv:2303.14123},
  year={2023}
}

@article{chen2022vision,
  title={Vision transformer adapter for dense predictions},
  author={Chen, Zhe and Duan, Yuchen and Wang, Wenhai and He, Junjun and Lu, Tong and Dai, Jifeng and Qiao, Yu},
  journal={arXiv preprint arXiv:2205.08534},
  year={2022}
}

\end{document}